\documentclass{article}

\usepackage{arxiv}

\usepackage[utf8]{inputenc} 
\usepackage[T1]{fontenc}    
\usepackage{hyperref}       
\usepackage{url}            
\usepackage{booktabs}       
\usepackage{amsfonts}       
\usepackage{nicefrac}       
\usepackage{microtype}      
\usepackage{graphicx}
\usepackage[numbers]{natbib}
\usepackage{doi}
\usepackage{amsmath}
\usepackage{listings}
\usepackage{threeparttable}
\usepackage{arydshln}
\usepackage[draft]{minted} 

\newrobustcmd{\B}{\bfseries}

\def\code#1{\texttt{#1}}

\title{MolGraph: a Python package for the implementation of molecular graphs and graph neural networks with TensorFlow and Keras}

\date{}

\author{
\href{https://orcid.org/0000-0002-5295-010X}{Alexander Kensert\textsuperscript{1,2,*}\includegraphics[scale=0.06]{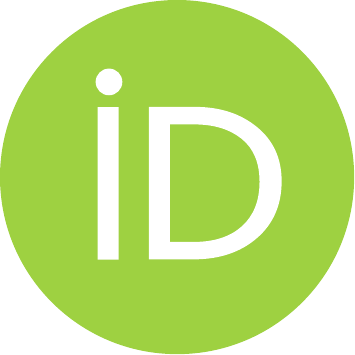}}, \href{https://orcid.org/0000-0001-8781-7184}{Gert Desmet\textsuperscript{2}\includegraphics[scale=0.06]{orcid.pdf}} and \href{https://orcid.org/0000-0001-5502-5801}{Deirdre Cabooter\textsuperscript{1}\includegraphics[scale=0.06]{orcid.pdf}}
} 

\renewcommand{\headeright}{A Preprint}
\renewcommand{\undertitle}{A Preprint}
\renewcommand{\shorttitle}{MolGraph}

\begin{document}

\maketitle

\begin{center}
\parbox{128mm}{
    \centering
    (1) University of Leuven (KU Leuven), Department for Pharmaceutical and Pharmacological
    Sciences, Pharmaceutical Analysis, Herestraat 49, 3000 Leuven, Belgium\vspace{1mm} \\
    (2) Vrije Universiteit Brussel (VUB), Department of Chemical Engineering, Pleinlaan 2,
    1050 Brussel, Belgium\vspace{1mm} \\
    (*) Corresponding author, email: \href{mailto:alexander.kensert@gmail.com}{alexander.kensert@gmail.com}\vspace{1mm}\\
    \vspace{10mm}
}
\end{center}

\begin{abstract}
    Molecular machine learning (ML) has proven important for tackling various molecular problems, such as predicting molecular properties based on molecular descriptors or fingerprints. Since relatively recently, graph neural network (GNN) algorithms have been implemented for molecular ML, showing comparable or superior performance to descriptor or fingerprint-based approaches. Although various tools and packages exist to apply GNNs in molecular ML, a new GNN package, named MolGraph, was developed in this work with the motivation to create GNN model pipelines highly compatible with the TensorFlow and Keras application programming interface (API). MolGraph also implements a chemistry module to accommodate the generation of small molecular graphs, which can be passed to a GNN algorithm to solve a molecular ML problem. To validate the GNNs, they were benchmarked against the datasets of MoleculeNet, as well as three chromatographic retention time datasets. The results on these benchmarks illustrate that the GNNs performed as expected. Additionally, the GNNs proved useful for molecular identification and improved interpretability of chromatographic retention time data. MolGraph is available at \url{https://github.com/akensert/molgraph}. Installation, tutorials and implementation details can be found at \url{https://molgraph.readthedocs.io/en/latest/}.
\end{abstract}
\vspace{3mm}


\section{Introduction}\label{sec:1}

\begin{figure}[ht]
      \centering
      \includegraphics[width=0.9\textwidth]{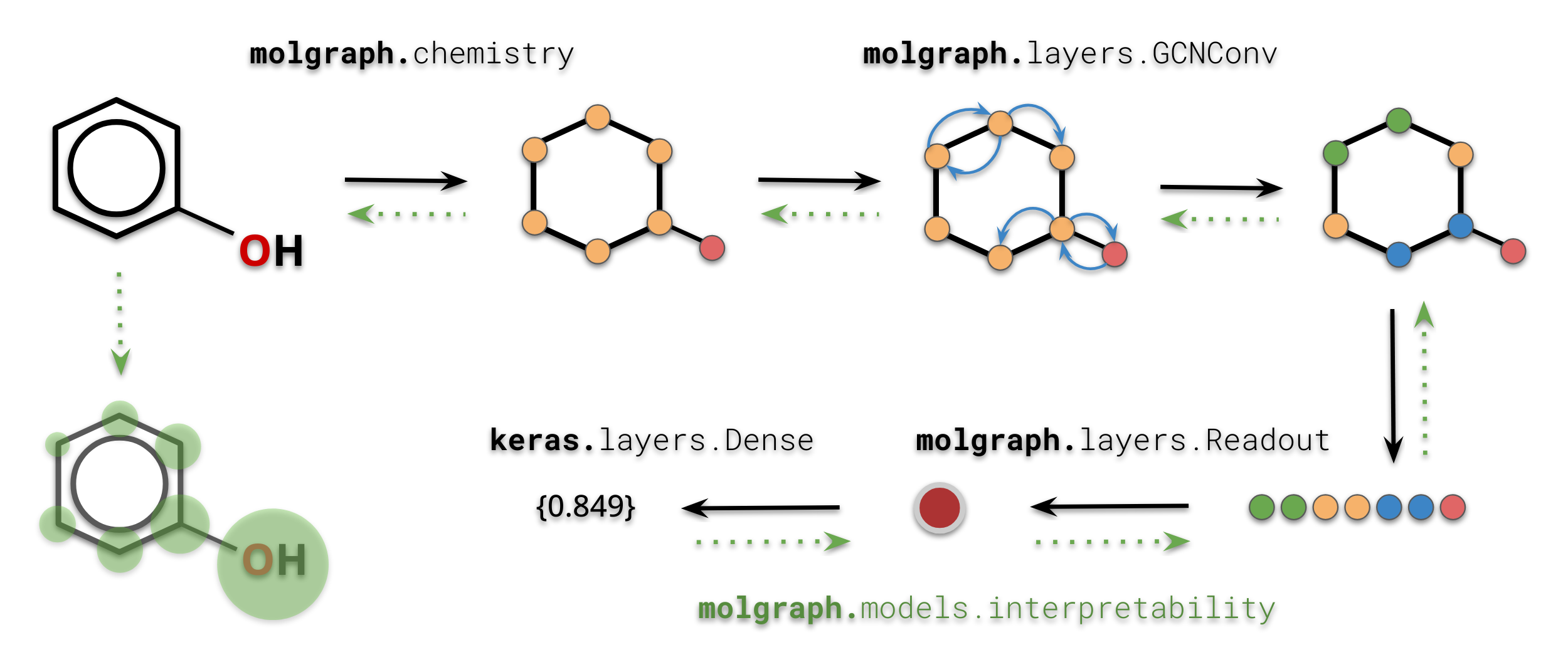}
      \caption{Schematic visualization of the different modules of MolGraph, and how they are utilized for molecular ML.}
      \label{fig:fig1}
    \end{figure}

    Molecular machine learning (ML) has played an important role in solving various molecular problems; including prediction of aqueous solubility\cite{intro1}, binding of drugs to specific target proteins\cite{intro2} and blood brain-barrier permeability\cite{intro3}. Specifically, molecular ML tackles these problems by linking structural information (e.g. molecular descriptors\cite{intro4} or fingerprints\cite{intro5}) to a desired output (e.g., aqueous solubility measurements, binding or no binding to a protein, or permeability). As molecules are naturally represented as graphs $G = (V,E)$, where $V$ is a set of vertices or nodes corresponding to the atoms, and $E$ a set of edges corresponding to the bonds, it is reasonable to encode molecules as graphs. By then subjecting these molecular graphs to graph neural network\cite{intro6, intro7} (GNN) algorithms, highly expressive and meaningful numerical representations of molecules for downstream classification or regression tasks can be obtained. As the application of GNNs for molecular ML has already proven useful for both predictive modeling\cite{intro6, intro7, intro8, intro9, intro10} and interpretability\cite{intro9, intro11}, the aim of this work was to develop a Python package, MolGraph (available at \url{https://github.com/akensert/molgraph}), specializing in molecular graphs and in building GNNs (see Figure \ref{fig:fig1}). Importantly, in addition to being built with TensorFlow\cite{intro12} (TF), MolGraph is also built to be highly compatible with TF's application programming interfaces (APIs), such as the Keras\cite{intro13} API, \code{tf.data.Dataset} API and \code{tf.saved\_model} API.  
    
    Many ML libraries exist in Python, these include Scikit-learn (for implementing classical ML models such as random forest, gradient boosting, multi-layer perceptron, support vector machines, etc.), and TF, Keras and Pytorch\cite{intro14} (for implementing deep neural networks). Built on top of these libraries, Python packages have been developed to fit certain applications. For instance, for GNN applications, these include Spektral\cite{intro15} (based on Keras/TF), Deep Graph Library\cite{intro16} (based on PyTorch, TF and Apache MXNet), DeepChem\cite{intro17} (based on Keras/TF and PyTorch), PyTorch Geometric\cite{intro18}, and Graph Nets\cite{intro19} (based on TF and Sonnet). Many of the exisiting packages for GNNs, such as DeepChem (a prominent Python package for molecular ML) and Spektral (a prominent Python package for GNN applications), have a broader focus than MolGraph and thus a larger set of tools and utilities. However, the significant advantage of MolGraph, is that it is implemented to be highly compatible with TF's APIs, making GNN implementations highly flexible, optimized and easy. The key feature of MolGraph that makes this possible is the implementation of a custom \code{GraphTensor}, which is implemented with TF's relatively new \code{tf.ExtensionType} API and thus works seamlessly with TF on every level; including \code{keras.Sequential} and \code{keras.Functional} model API, \code{tf.saved\_model} API and \code{tf.data.Dataset} API.

\section{API}\label{sec:2}

    \subsection{The graph}
        
        In contrast to text or image data, which is often encoded as a single floating point tensor, graph data requires multiple separate tensors to represent it. Hence, to improve model building using graphs as input, a \textit{composite} tensor, namely a \code{GraphTensor} class, was implemented. This composite tensor incorporates multiple fields of graph data into a single entity:
    
        \begin{itemize}

            \item \code{sizes} (required); the sizes of the encoded (sub)graphs.
            
            \item \code{node\_feature} (required); the node (atom) features. Row $i$ corresponds to node $i$.
            
            \item \code{edge\_src} (required); the source nodes of the edges (bonds). Entry $i$ corresponds to node $i$.
            
            \item \code{edge\_dst} (required); the destination nodes of the edges (bonds). Entry $i$ corresponds to node $i$.

            \item \code{edge\_feature} (optional); the edge (bond) features. Row $j$ corresponds to edge $j$.

            \item \code{edge\_weight} (optional); the weights assigned to each edge. Row $j$ corresponds to edge $j$. 
            
            \item \code{node\_position} (optional); the node positional encoding. Row $i$ corresponds to node $i$.
            
        \end{itemize}
        
        \noindent Each of the listed fields of data can be accessed as attributes from a \code{GraphTensor} instance; and each plays an important role in producing a final molecular embedding for downstream regression or classification tasks, such as solubility or ligand-protein interaction prediction, respectively. Notably, although molecular graphs are usually \emph{undirected} graphs, both source and destination nodes are encoded separately (in \code{edge\_src} and \code{edge\_dst} respectively); which means that the \code{GraphTensor} supports \emph{directed} graphs as well. 
    
        As required by ML applications, a single \code{GraphTensor} instance needs to be able to encode multiple graphs (e.g. a batch of molecular graphs). To efficiently encode and transform a batch of molecular graphs, the \code{GraphTensor} instance encodes these molecular graphs as a single disjoint graph $G=G_{1}\cup G_{2} \; ... \cup G_{n}$, where $G_{i}$ is the i:th molecular graph. Although mathematically equivalent, a single disjoint graph can be operated on more efficiently than multiple separate subgraphs. However, if desired, the \code{GraphTensor} instance implements methods to go between the two  states:
        
        \begin{itemize}
            \item \code{separate()}, which "unpacks" the disjoint graph into its subgraphs $G=\{G_{1}, G_{2},..., G_{n}\}$.
            \item \code{merge()}, which "packs" the subgraphs into a disjoint graph $G=G_{1}\cup G_{2} \; ...  \cup G_{n}$.
        \end{itemize}
        
        \noindent Furthermore, the \code{GraphTensor} implements methods to update, remove or add fields of data:
        
        \begin{itemize}
            \item \code{update()}, which adds new data to the graph, either an existing or non-existing field.
            \item \code{remove()}, which removes data from the graph.
        \end{itemize}

        \noindent  Finally, the \code{GraphTensor} instance is not only implemented to encode graphs, but also to conveniently propagate information within the graph:

        \begin{itemize}
            \item \code{propagate()}, which aggregates, for each node in the graph, information from adjacent nodes.
        \end{itemize}

        \noindent Although many attributes and methods exist for a \code{GraphTensor} instance, the user rarely needs to keep track of or think about how these different parts interact and how the graph propagates within GNN \code{layers}. Most often, the user simply needs to (1) initialize a \code{GraphTensor} instance (performed by MolGraph's \code{chemistry} module), (2) pass it to a \code{tf.data.Dataset} and (3) finally a \code{keras.Sequential} model comprising GNN \code{layers} (from \code{molgraph.layers} module). In other words, batching and propagation occur automatically in the \code{tf.data.Dataset} API and MolGraph's \code{layers} API respectively.

    \subsection{The molecular graph}
        The initial state of a single or a batch of molecular graph(s), encoded as a single \code{GraphTensor} instance, is produced by the \code{chemistry} module of MolGraph. Specifically, a \code{MolecularGraphEncoder} transforms string representations of molecules (e.g., SMILES or InChI) into a molecular graph encoded as a \code{GraphTensor}. \code{edge\_src} and \code{edge\_dst} are computed internally based on the adjacency matrix corresponding to the molecular graph, and \code{node\_feature} and \code{edge\_feature} are both computed from either a \code{Featurizer} or \code{Tokenizer} based on a list of \code{features}. There are a number of different \code{features} that can be computed with MolGraph (see Table \ref{tab:table1}), each of which aims to provide some useful information about the nodes and edges of the molecular graph.
        
        As atoms of a molecule may be assigned 3D coordinates, MolGraph also supplies a \code{MolecularGraphEncoder3D} to compute the distance geometry of a molecule, in addition to the required data. Specialized GNNs, such as DTNN\cite{other20}, can then exploit such information to generate a potentially improved representation of the molecule for downstream regression or classification tasks.

    \begin{table}
     \caption{List of available atom and bond features of MolGraph}
     \label{tab:table1}
      \centering
        \begin{tabular}{llll}
        \toprule
         \multicolumn{2}{c}{Atom features} & \multicolumn{2}{c}{Bond features}  \\
         \midrule
            Description & Encoding & Description & Encoding \\
          \midrule 
            Chiral center & Binary & Bond type & One-hot \\
            Chirality (R, S or None) & One-hot & Conjugated & Binary  \\
            Crippen Log P contribution & Float & Rotatable & Binary \\
            Crippen Molar refractivity contribution & Float & Stereo & One-hot  \\
            Degree & One-hot & Part of ring & Binary \\
            Formal charge & One-hot & Part of ring of size \emph{n} & One-hot \\
            Gasteiger charge & Float & & \\
            Hybridization & One-hot & & \\
            Aromatic & Binary & & \\
            Hydrogen donor & Binary & & \\
            Hydrogen acceptor & Binary & & \\
            Hetero & Binary & & \\
            Part of ring & Binary & & \\
            Part of ring of size \emph{n} & One-hot & & \\
            Labute accessible surface area contribution & Float & & \\
            Number of hydrogens & One-hot & & \\
            Number of radical electrons & One-hot & & \\
            Number of valence electrons & One-hot & & \\
            Symbol (atom type) & One-hot & & \\
            Topological polar surface area contribution & Float & & \\
        \bottomrule
      \end{tabular}
    \end{table}

    \subsection{TF records}
        To support large graphs, i.e. a large dataset of molecules (exceeding 50,000 in numbers), which usually does not fit into (random access) memory of a personal computer, a TF Records module (\code{molgraph.chemistry.tf\_records}) is implemented in MolGraph. In brief, TF records allow the user to save data to disk, which can then efficiently be read (from disk) for modeling. Hence, the user may write data, namely molecular graphs and associated labels, as TF records, which can then be read in a computationally and memory efficient way later on. 
        
    \subsection{GNN layers}
        Similar to how a feed-forward neural network (FFNN) is usually a stack of fully-connected layers, a GNN is usually a stack of GNN layers. Although the GNN layers of MolGraph take many shapes and forms, they are readily interchangeable, making it easy to replace one GNN layer with another. These GNN layers simply take as input a \code{GraphTensor} instance, and outputs a new \code{GraphTensor} instance with an updated set of node (and possibly edge) features.
        
\begin{figure}[t]
          \centering
          \includegraphics[width=0.95\textwidth]{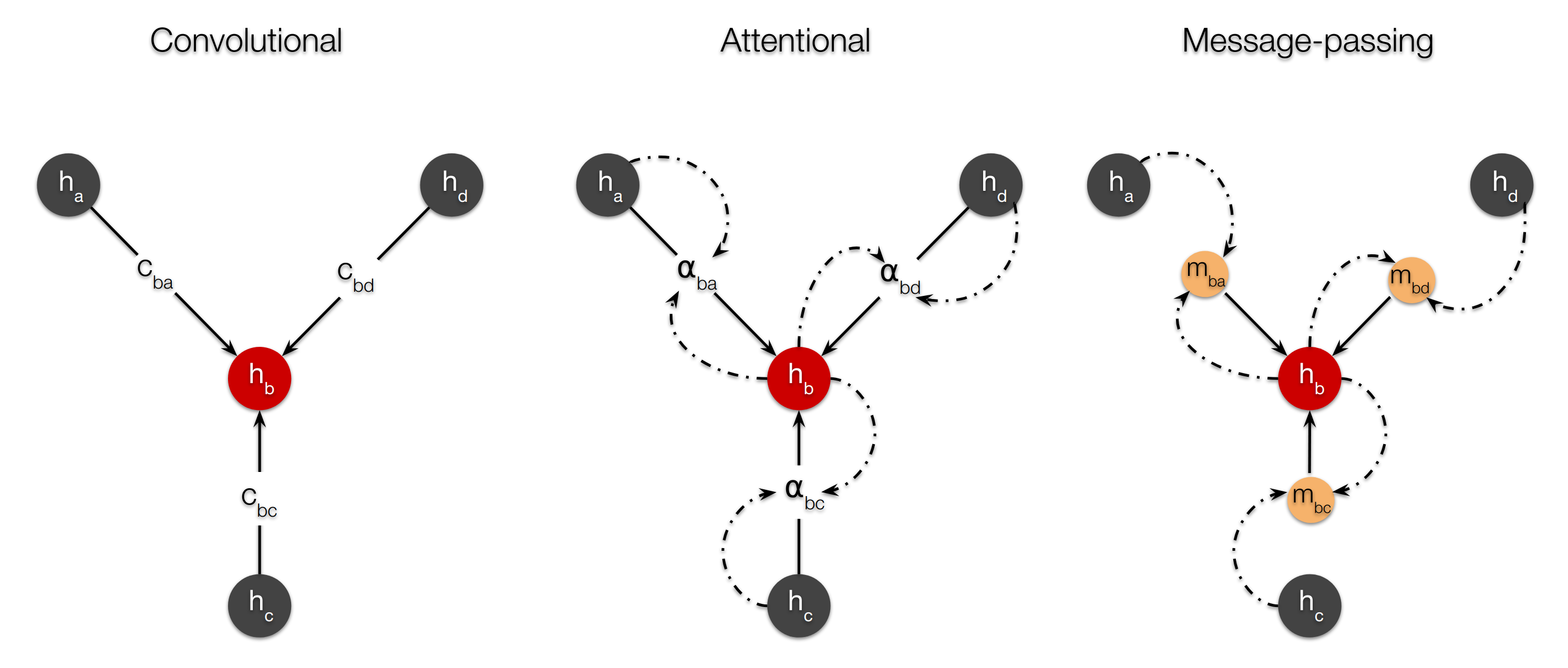}
          \caption{Schematic illustration of the three main types of graph neural networks.}
          \label{fig:fig2}
        \end{figure}

        Specifically, a GNN layer operates in two steps: (1) transformation of the node features and (2) aggregation of the node features. In the simplest case, the transformation is a single linear transformation via a learnable weight matrix on the node features, directly followed by an aggregation, by averaging all source nodes' features for each destination node. For some of the GNN layers, however, these two steps are more complicated and expressive, wherein the updated node features might depend on the relevant edge features and/or an attention mechanism.
        The GNN layers of MolGraph can be separated into three different categories: convolutional (e.g., GCN\cite{other21, other22} and GIN\cite{other22, other23}), attentional (e.g., GAT\cite{other22, other24} and GMM\cite{other22, other25}) and message passing (e.g., MPNN\cite{intro8}) (see Figure \ref{fig:fig2}). 
        
        The convolutional layer performs an arbitrary transformation on the neighboring node features $h_{j}$ of node $i$, and subsequently aggregates them via normalization coefficients $c_{ij}$, as follows:
        
        \begin{equation}
            h_{i}^{(l+1)} = \sigma\left(\sum_{j \in N(i)}{c_{ij}\psi(h_{j}^{(l)})}\right),
        \end{equation}
        
        were $\sigma$ is a non-linear activation function, e.g., \code{ReLU}. The attentional layer adds an attention mechanism, as a replacement for $c_{ij}$:
        
        \begin{equation}
            h_{i}^{(l+1)} = \sigma\left(\sum_{j \in N(i)}{\alpha_{ij}^{(l)}\psi(h_{j}^{(l)})}\right),
        \end{equation}
        
        where $\alpha$ denotes the attention coefficients, computed based on $h_{i}^{(l)}$ and $h_{j}^{(l)}$:
        
        \begin{equation}
            \alpha_{ij}^{(l)} = a(h_{i}^{(l)}, h_{j}^{(l)}),
        \end{equation}
        
        and $a$ is a function which computes the attention coefficients $\alpha_{ij}^{(l)}$. The message-passing layer uses a so-called message-function which computes the “message” to be propagated to the destination nodes. In comparison with the attentional (and convolutional) layer, the message-passing layer does not pass $h_{j}^{(l)}$ directly, but indirectly, through the message-function; which computes the message based on $h_{i}^{(l)}$ and $h_{j}^{(l)}$:
        
        \begin{equation}
            h_{i}^{(l+1)} = \sigma\left(\sum_{j \in N(i)}{\psi(h_{i}^{(l)}, h_{j}^{(l)})}\right)
        \end{equation}
        
        As can be seen, the transformation function depends both on $h_{i}^{(l)}$ and $h_{j}^{(l)}$, and not only $h_{j}^{(l)}$, which makes this approach the most expressive.
        
        In addition to the above, a given GNN layer may implement an aggregation step based on edge features $e_{ij}$, in addition to node features; where $e_{ij}$ indicates the edge features of the edge between node $i$ and $j$. For example, the graph attention network (GAT) of MolGraph optionally incorporates edge features, wherein the attention coefficients $\alpha_{ij}^{(l)}$ are computed based on both the node features and the associated edge features:
        
        \begin{equation}
            \alpha_{ij}^{(l)} = a(h_{i}^{(l)}, h_{j}^{(l)}, e_{ij}^{(l)})
        \end{equation}
    
    \subsection{Readout}
    
        After $L$ steps of transformation and aggregation, $h_{i}^{(L)}$ is obtained for all nodes $i$ in the graph. $h_{i}^{(L)}$ can be used to perform e.g., node predictions, edge predictions or graph predictions. For graph predictions, $h_{i}^{(L)} \; \forall i \in N(G_{k})$ (node embeddings of graph $k$) needs to be reduced to $z_{k}$ (graph embedding of graph $k$). MolGraph implements several different layers that perform this reduction (known as \emph{readout}). The most basic readout layer simply sums over the nodes as follows:
        
        \begin{equation}
            z_{k} = \sum_{i \in N(G_{k})}{h_{i}^{(L)}}
        \end{equation}
        
        Subsequently, $z_{k}$ can be passed to an FFNN (denoted $f_{\theta}$) to get a prediction (denoted $\hat{y}_{k}$):
        
        \begin{equation}
            \hat{y}_{k} = f_{\theta}(z_{k})
        \end{equation}

    \subsection{GNN models}

        \paragraph{Pretraining}
            Pretraining models to improve downstream modeling has proven important for many applications, including natural language processing and image recognition \cite{other36, other37}. For this reason, MolGraph implements a few pretraining strategies, including masked graph modeling. Inspired by masked language modeling \cite{other36} wherein words in a piece of text are randomly masked out and predicted, masked graph modeling masks out nodes (atoms) and edges (bonds) to be predicted. By performing this self-supervising task on millions of examples (molecules), it is hoped that the underlying GNN model has acquired a "good understanding" of what graphs (molecules) are and the interactions within, consequently improving downstream modeling such as molecular property predictions.
            
        \paragraph{Classification and regression}
            Building a model with Keras is convenient, allowing users to build complex deep neural networks to solve problems in a couple of lines of code. It was therefore imperative to implement a custom \code{GraphTensor} class to allow MolGraph to utilize Keras to its full capacity. Specifically, MolGraph's GNN layers can be passed to \code{keras.Sequential} API, used in \code{keras.Functional} API or used with Keras subclassed models (\code{keras.Model}); and methods such as \code{fit}, \code{evaluate} and \code{predict} work exactly as expected, accepting either a \code{GraphTensor} instance (which will internally be mini-batched) or a \code{tf.data.Dataset} constructed from a \code{GraphTensor} instance. 
    
        \paragraph{Saliency and gradient activation mapping}
            There is an increasing interest in better understanding and interpreting deep learning models (including GNNs). Among many techniques that can be used for this purpose, two prominent and well researched techniques are saliency and gradient activation mapping\cite{intro11}. To allow better understanding and interpretability of the GNNs, both saliency and gradient activation mapping modules are implemented in MolGraph (\code{molgraph.models.interpretability}). These two techniques can be used to better understand what molecular substructures are important for the prediction; in other words, what parts of the molecules the GNN is “looking” at. 

\section{Experiments and results}\label{sec:3}

    \subsection{Molecular property predictions} 
    
        To validate the performance of MolGraph's GNN \code{layers} to predict properties of molecules, they were implemented with \code{keras.Sequential} model API, and subsequently fitted to and evaluated on a selection of benchmarks. The benchmarks included 15 datasets from MoleculeNet\cite{other26}, one dataset from Domingo-Almenara et al.\cite{other27} (in this study, referred to as “SMRT”; an abbreviation for \emph{small molecular retention times}), and two datasets from Bonini et al.\cite{other28} (in this study, referred to as “RPLC” and “HILIC”; reflecting which chromatographic separation mode was used, namely, \emph{reversed-phase liquid chromatography} and \emph{hydrophilic interaction liquid chromatography}). In addition to the inclusion of the aforementioned benchmarks, accompanying model results from the corresponding studies were also included as reference points. Note that, although both MoleculeNet's and MolGraph's MPNN and DTNN models are based on the original implementations \cite{other20, intro8}, there are differences between the implementations. For instance, MolGraph's GNN \code{layers} apply post-processing steps such as batch normalization and skip connection; and furthermore, no distance geometric information is used in MolGraph's MPNN model.

        As the literature models merely serve as reference points for performance, there were no attempts in systematically tuning the hyperparameters to improve performance. Instead, the default hyperparameters were used for each layer of each model. Furthermore, for the same reason, no replicate runs were performed. As for the optimization procedure, the Adam\cite{other29} optimizer was used, with a starting learning rate of 0.0001 and ending learning rate of 0.000001, which decayed, on every plateau (defined as no improvement for 10 epochs), by a factor of 0.1. The training stopped when no improvement was made for 20 epochs, and the weights of the best epoch were restored for the test sets. The loss functions used were, for classification, binary cross entropy (BCE); and for regression, depending on the evaluation metric, mean absolute error (MAE), root mean squared error (RMSE) or Huber.

        Based on the predictive performance (Table \ref{tab:table2} and \ref{tab:table3}), the GNN models of MolGraph performed well. Namely, for the regression datasets (Table \ref{tab:table2}), almost all of MolGraph's models had a lower RMSE or mean relative error (MRE) compared to the literature models. For the classification datasets (Table \ref{tab:table3}), most of MolGraph's models had a similar ($\pm 0.05$ AUC) or higher AUC compared to the literature models. The exceptions were \emph{qm7} and \emph{muv}, where MoleculeNet's \emph{DTNN} and \emph{Weave} had a noticeably lower and higher score respectively. 
        
        Furthermore, the results illustrate that attentional and message-passing type models perform marginally better than the convolutional type models; suggesting that increasing the complexity of the models could improve the predictive performance --- though at the cost of significantly increasing the computational cost. On the other hand, the results do not reveal improvements in predictive performance when incorporating edge features. 
        
        Finally, the results also illustrate that models, such as the graph transformer (GT), performed markedly well on some datasets compared to other datasets; suggesting a strong dataset dependence. For GT specifically, the low RMSE on the quantum-chemistry datasets (and on the other hand low AUC on the physiology datasets), could be explained by the relatively complex attention mechanism in the GT \code{layers}, which favors smaller molecular graphs (those of the quantum-chemistry datasets) and disfavors larger molecular graphs (those of the physiology datasets). It is left for future studies to further investigate the predictive performance of the different GNN models of MolGraph, and how they compare with the current state-of-the-art.
        
        \begin{table}[ht]
            \caption{
                Predictive performance on the three quantum-chemistry datasets, three physico-chemistry datasets and five chromatographic retention time datasets. The dashed line visually separates the models (and results) of this study (MolGraph) from the models (and results) of the other studies (see footnote of table). “(E)” indicates that edge features were used. Each value associated with the quantum-chemistry datasets indicates the mean absolute error (MAE); each value associated with the physico-chemistry datasets indicates the root mean squared error (RMSE); and each value associated with the chromatographic RT datasets indicate the mean relative error (MRE). Convolutional type models: GCN, RGCN(E), GIN, and GraphSage; attentional type models: GAT, GAT(E), GatedGCN, GatedGCN(E), GMM, GT, and GT(E); and message-passing type models: MPNN(E).
            }
            \label{tab:table2}
            \centering
            \begin{tabular}{lccccccccccc}
                \toprule
                & \multicolumn{3}{c}{quantum-chemistry} & \multicolumn{3}{c}{physico-chemistry} & \multicolumn{5}{c}{chromatographic RT}  \\
                \midrule
                & qm7     & qm8    & qm9    & esol   & lipoph.  & freesolv & SMRT & \multicolumn{2}{c}{RPLC\footnotemark[4]} & \multicolumn{2}{c}{HILIC\footnotemark[4]}\\
                &      &     &     &    &  &  &  & test & ext. & test & ext. \\
                \midrule
                GCN         & 18.9011 & 0.0086 & 2.4795 & 0.5682 & 0.5412 &\B0.8453 & 0.038 & 0.09 &\B0.13 & 0.16 & 0.35 \\
                RGCN(E)     & 13.6039 & 0.0086 & 2.0556 & 0.5992 & 0.5639 & 1.3143 & 0.037 & 0.11 & 0.19 & 0.18 & 0.33 \\
                GIN         & 19.2505 &\B0.0084 & 2.0315 & 0.5865 & 0.5381 & 0.9066  & 0.037 & 0.10 & 0.17 & 0.17 & 0.30 \\
                GraphSage   & 20.4176 & 0.0092 & 2.4739 & 0.5919 & 0.5449 & 1.6146 & 0.039 & 0.10 & 0.15 & 0.17 & 0.32 \\
                GAT         & 22.0590 & 0.0088 & 2.6678 & 0.6471 & 0.5380 & 1.0530 & 0.038 & 0.09 & 0.15 & 0.18 &\B0.28 \\
                GAT(E)      & 17.9556 & 0.0087 & 2.1902 & 0.5961 & 0.5402 & 1.1959 & 0.038 & 0.11 & 0.16 & 0.16 & 0.29 \\
                GatedGCN    & 12.2210 & 0.0088 & 2.3740 & 0.5914 &\B0.5253 & 1.1264 & 0.037 & 0.10 & 0.15 & 0.16 & 0.31 \\ 
                GatedGCN(E) & 10.2520 & 0.008 & 1.9858 & 0.5755 & 0.5422 & 1.4468 &\B0.037 & 0.10 & 0.14 & 0.16 & 0.32 \\
                GMM         & 18.5327 & 0.0091 & 2.4415 & \B 0.5453 & 0.5417 & 0.8613 & 0.038 &\B0.09 & 0.15 & 0.16 & 0.28 \\
                GT          & 11.1366 & 0.0101 & 2.2205 & 0.6713 & 0.6195 & 1.0952 & & & & & \\
                GT(E)       &\B7.5603 & 0.0095 & 1.7375 & 0.6992 & 0.5703 & 1.1807 & & & & & \\
                MPNN(E)     & 15.0094 & 0.0089 & 2.5013 & 0.5505 & 0.5365 & 0.9538  & 0.038 & 0.09 & 0.16 &\B0.15 & 0.32 \\
                DTNN        & 10.8403 & 0.0114 &\B1.3970 & & & & & & & & \\
                \hdashline
                GC\footnotemark[1]   & 77.90 & 0.0148 & 4.7 & 0.97 & 0.655 & 1.40 & & & & & \\
                MPNN\footnotemark[1] & & 0.0143 & 3.2 & 0.58 & 0.719 & 1.15 & & & & & \\
                DTNN\footnotemark[1] & 8.8 & 0.0169 & 2.4 & & & & & & & & \\
                DLM\footnotemark[2]  & & & & & &  & 0.068 & & & & \\
                Keras\footnotemark[3] & & & & & & & & 0.13 & 0.18 & 0.2 & 0.46 \\ 
                \bottomrule
            \end{tabular}
            \footnotetext[1]{
                Values extracted from the MoleculeNet paper\cite{other26}. The exact dataset splits producing these results may differ from the dataset splits of this study.
            }
            \footnotetext[2]{
                Values extracted from Domingo-Almenara et al.\cite{other27} The exact dataset splits producing these results differ from the dataset splits of this study. In \cite{other27}, a training and test set were used, with 75\% and 25\% of the data, respectively. In this study, a training, validation and test set were used, with 70\%, 5\% and 25\% of the data, respectively.
            }
            \footnotetext[3]{
                Values extracted from Bonini et al.\cite{other28}. "Keras" refers to the name of their model; a deep fully-connected neural network, implemented in Keras. 
            }
            \footnotetext[4]{
                These datasets include two test sets: a "test set" and an "external set". See Bonini et al.\cite{other28} for more information.
            }
        \end{table}

        \clearpage
        
        \begin{table}[ht]
            \caption{
                Predictive performance on the four biophysics datasets and five physiology datasets. The dashed line visually separates the models (and results) of this study (MolGraph) from the models (and results) of the other studies (see footnote of table). “(E)” indicates that edge features were used. Each value associated with \emph{muv} and \emph{pcba} indicates the area under the precision-recall curve (PRC-AUC). Each remaining value indicates the area under the receiver operating characteristic curve (ROC-AUC). Convolutional type models: GCN, RGCN(E), GIN, and GraphSage; attentional type models: GAT, GAT(E), GatedGCN, GatedGCN(E), GMM, GT, and GT(E); and message-passing type models: MPNN(E).
            }
            \label{tab:table3}
            \centering
            \begin{tabular}{lccccccccc}
                \toprule
                    & \multicolumn{4}{c}{biophysics} & \multicolumn{5}{c}{physiology}  \\
                \midrule
                    & muv & hiv & pcba & bace & clintox & sider & toxcast & tox21 & bbbp \\
                \midrule
                GCN         & 0.0373 & 0.7964 & 0.1387 & 0.8111 & 0.7800 & 0.6230 & 0.7472 & 0.8233 & 0.6872 \\
                RGCN(E)     & 0.0377 & 0.7945 & 0.1478 & 0.8213 & 0.7268 &\B0.6583 & 0.7501 & 0.8209 & 0.6948 \\
                GIN         & 0.0033 & 0.7677 &\B0.1524 & 0.8231 & 0.8188 & 0.6202 & 0.7447 & \B 0.8420 & 0.6763 \\ 
                GraphSage   & 0.0331 & 0.7763 & 0.1345 & 0.8129 & 0.8320 & 0.6376 & 0.7357 & 0.8264 & 0.6969 \\ 
                GAT         & 0.0308 &\B0.8009 & 0.1450 & 0.8294 & 0.7515 & 0.6091 & 0.7413 & 0.8336 &\B0.7196 \\ 
                GAT(E)      & 0.0279 & 0.7756 & 0.1500 & 0.7806 & 0.7483 & 0.6204 & 0.7342 & 0.8151 & 0.6938 \\
                GatedGCN    & 0.0693 & 0.7669 & 0.1475 & 0.8063 & 0.7754 & 0.5995 & 0.7397 & 0.8348 & 0.6807 \\
                GatedGCN(E) & 0.0160 & 0.7689 & 0.1482 & 0.7974 &\B0.8746 & 0.6302 & 0.7384 & 0.8287 & 0.6967 \\
                GMM         & 0.0380 & 0.7643 & 0.1466 & 0.7732 & 0.8128 & 0.6280 & 0.7473 & 0.8283 & 0.6903 \\
                GT          & 0.0034 & 0.7493 & 0.1356 & 0.7325 & 0.8322 & 0.5704 & 0.7309 & 0.8057 & 0.6857 \\
                GT(E)       & 0.0145 & 0.7543 & 0.1404 & 0.7958 & 0.8430 & 0.5922 & 0.7264 & 0.7982 & 0.6726 \\ 
                MPNN(E)     & 0.0400 & 0.7603 & 0.1427 & \B 0.8320 & 0.8329 & 0.6400 &\B0.7524 & 0.8345 & 0.6942 \\ 
                \hdashline
                GC\footnotemark[1]    & 0.046 & 0.763 & 0.136 & 0.783 & 0.807 & 0.638 & 0.716 & 0.829 & 0.690 \\
                Weave\footnotemark[1] &\B0.109 & 0.703 &  & 0.806 & 0.832 & 0.581 & 0.742 & 0.820 & 0.671 \\
                \bottomrule
            \end{tabular}
            \footnotetext[1]{
                Values extracted from the MoleculeNet paper \cite{other26}. The exact dataset splits producing these results may differ from the dataset splits of this study.
            }
        \end{table}

    \clearpage

    \subsection{Molecular identification}
        In addition to benchmarking the predictive performance of the GNNs of MolGraph on the MoleculeNet datasets, it was also of interest to verify the capabilities of a graph convolutional network (GCN) to assist in molecular identification. For that purpose, a plant metabolite database (PLaSMA, available at \url{http://plasma.riken.jp/}) consisting of liquid chromatography-tandem mass spectrometry (LC-MS/MS) data was used. In brief, the identity (structure) of an unknown molecule (analyte) in a given sample can be predicted based on LC-MS/MS data; where the MS/MS data is used to predict the structure of the analyte (using a software such as MSFinder\cite{other30, other31}), and the LC RTs are used as an additional confirmation of the proposed structure. Thus, the GCN was trained to correlate the structure of known molecules in the PLaSMA dataset to their associated RTs, and subsequently used to obtain more confidence about proposed structures of analytes based on their experimental RT. The PLaSMA dataset was divided into training, validation and test sets; in which the training set was used to train the model; the validation set was used to supervise the training and to determine thresholds for the RT filter; and the test set was used to evaluate how well the GCN model could filter out structures (candidates) suggested by the MSFinder software. 
        
\begin{figure}[t]
            \centering
            \includegraphics[width=0.95\textwidth]{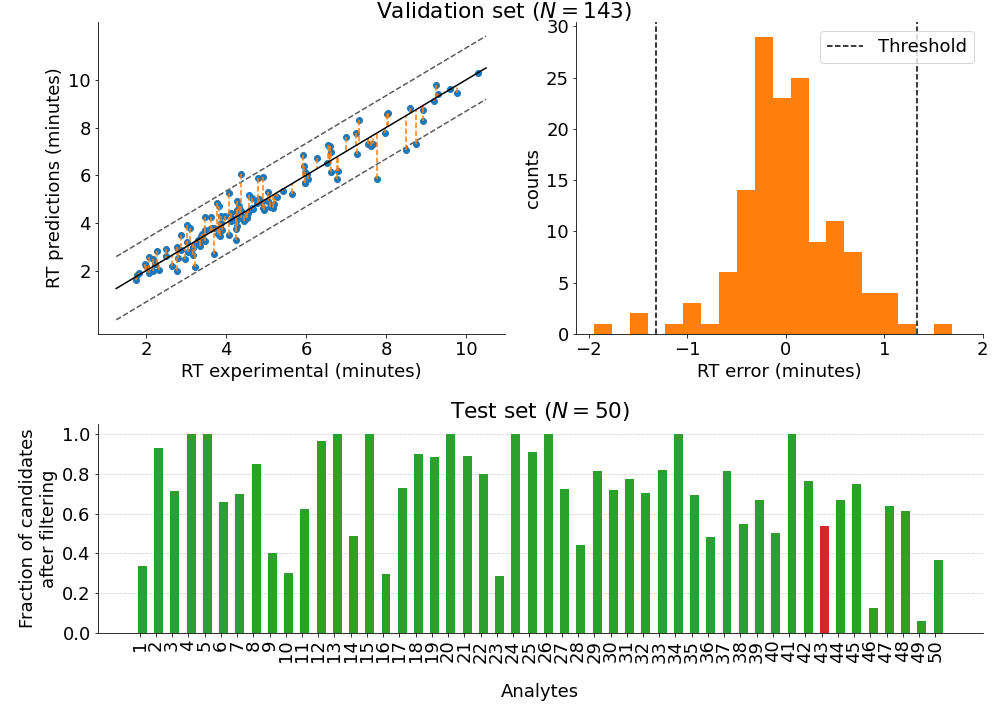}
            \caption{
                Retention time (RT) filter. Thresholds ($-1.313$ and $1.337$) were computed from the validation set and used to filter out candidates suggested by MSFinder. The red bar indicates a false negative.
            }
            \label{fig:fig3}
        \end{figure}

        Figure \ref{fig:fig3} aims to visualize the process of eliminating candidates proposed by MSFinder. The upper plots of Figure \ref{fig:fig3} illustrate how the validation set was used to define the RT filter (two \emph{thresholds}, or \emph{bounds}) that decided whether a candidate would be discarded or not as a possible identity of the analyte. Specifically, if the distance between the predicted RT of the candidate and the RT of the analyte was within the bounds, it was kept, otherwise, discarded. The bounds were defined by the mean and standard deviation of the errors, specifically, $\mu \pm 2.58 \sigma$ ($-1.313$ and $1.337$). The bottom subplot of Figure \ref{fig:fig3} illustrates to what extent the GCN could filter out candidates with the RT filter. As can be seen, a significant portion (about 40\%) of the candidates were filtered out, although in specific cases, none, or very few, got filtered out. In one case, a false negative occurred — i.e., the candidate matching the identity of the analyte was discarded. Overall, the results show evidence that the utilization of the GCN for RT predictions may facilitate the process of identifying analytes.

        To inspect the filtering procedure in more detail, the MSFinder candidate scores, as well as predicted candidate RTs (of the GCN) corresponding to analyte 10 (of Figure \ref{fig:fig3}), are shown in Table \ref{tab:table4}. The candidate in bold is the most likely identity based on the MSFinder score and RT filter. In this specific case, the predicted RT proved to be important for correctly identifying the analyte. Comparing MSFinder's three highest scoring candidates with the fourth highest scoring candidate, which ended up being the most likely candidate after applying the RT filter, it can be observed that the three highest scoring candidates are significantly more polar --- hence their low RTs in an RPLC setting. As the analyte had a significantly higher RT (because it was significantly less polar), the RT filter sufficed to discard them. Although this is just a single example, it aims to illustrate how combining information of predicted structures based on MS/MS data and predicted RTs can increase the chances of correctly identifying an analyte. 
        
        \begin{table}
            \caption{
                Analyte 10. In bold: the most likely identity of the analyte, based on both the MSFinder score and the difference between predicted RT and experimental RT.
            }
            \label{tab:table4}
            \centering
            \begin{tabular}{rcccc}
                \toprule
                \multicolumn{1}{l}{Target Identity:} & \multicolumn{4}{l}{CS(=O)CCCCCCCN=C=S} \\
                \multicolumn{1}{l}{Target RT:} & \multicolumn{4}{l}{5.41 (min)} \\ 
                \midrule
                \multicolumn{1}{c}{Candidates} & \multicolumn{1}{c}{\parbox{2cm}{\centering Score (MSFinder)}} &  \multicolumn{1}{c}{\parbox{2cm}{\centering Predicted candidate RT (min)}} & \multicolumn{1}{c}{\parbox{1.5cm}{\centering RT difference}} & \multicolumn{1}{c}{\parbox{1.5cm}{\centering Filtered out}} \\
                \midrule
                NC(CCOC(=O)CCC(=O)O)C(=O)O & 7.32 & 1.39 & 4.02 & Yes \\ 
                CC(=O)NC1C(=O)OC(CO)C(O)C1O & 7.12 & 1.71 & 3.70 & Yes \\ 
                CC(O)C(=O)NC(CCC(=O)O)C(=O)O & 7.06 & 2.03 & 3.38 & Yes \\
                \B CS(=O)CCCCCCCN=C=S & \B 7.06 & \B 6.07 & \B -0.66 & \B No \\
                CC(OC(=O)CCC(N)C(=O)O)C(=O)O & 6.87 & 1.63 & 3.78 & Yes \\
                Cc1ccc(C=NNc2cn[nH]c(=O)n2)o1 & 6.73 & 2.60 & 2.81 & Yes \\
                Cc1nonc1NC(=O)Nc1cccnc1 & 5.95 & 1.94 & 3.47 & Yes \\
                CSC=CC(=O)NC=Cc1ccccc1 & 5.78 & 4.63 & 0.78 & No \\
                CSC=CNC(=O)C=Cc1ccccc1 & 5.72 & 4.95 & 0.46 & No \\
                CCOC(=O)NP(=O)(N1CC1)N1CC1 & 5.34 & 2.12 & 3.29 & Yes \\
                \bottomrule
            \end{tabular}
        \end{table}

    \clearpage 

    \subsection{Gradient activation maps}
\begin{figure}[!t]
            \centering
            \includegraphics[width=0.75\textwidth]{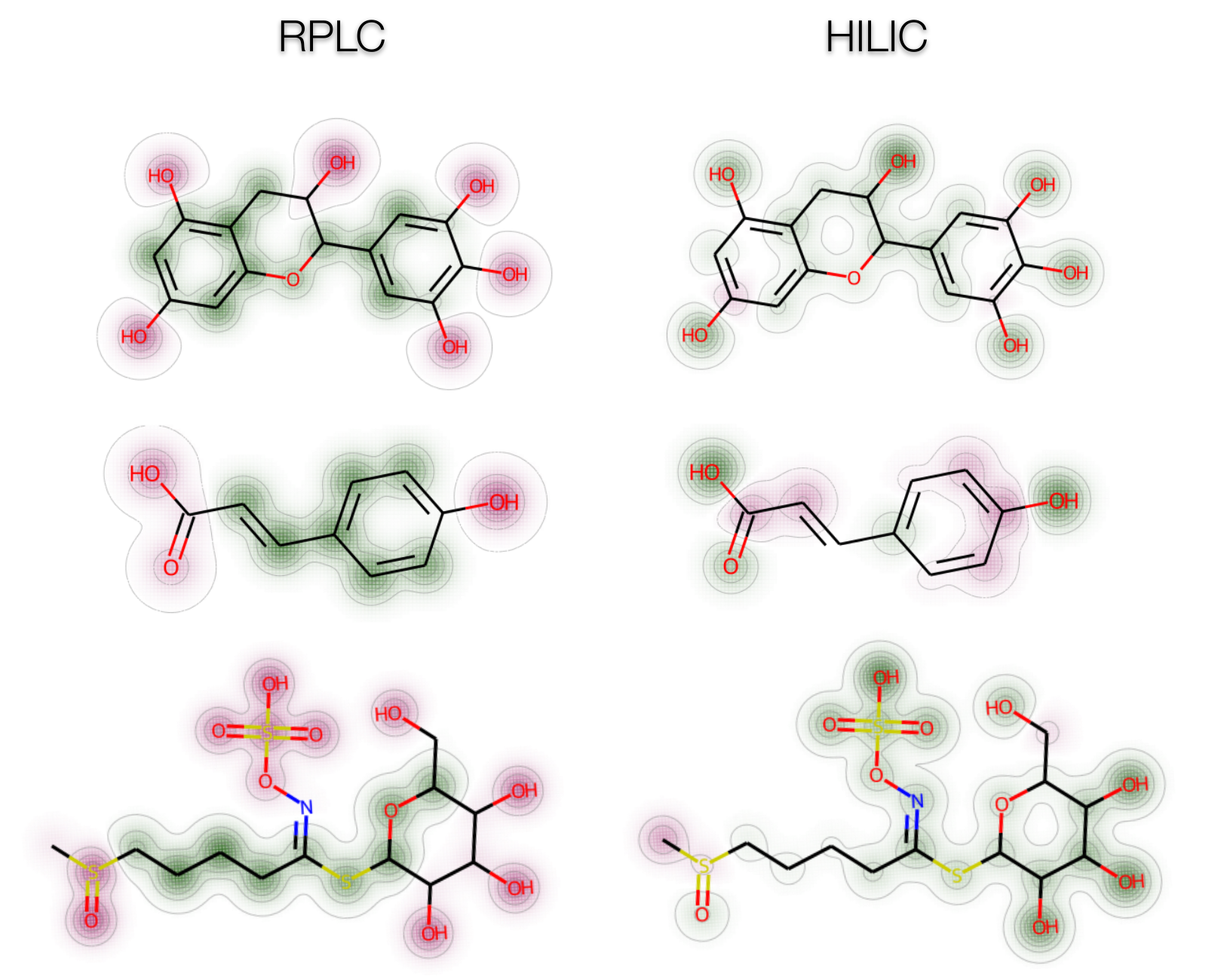}
            \caption{
                Comparison of gradient activation maps between the RPLC dataset and the HILIC dataset. Gradient activation maps (in green or purple) indicate how substructures contribute to retention (positively or negatively, respectively). The number of contour lines indicates to what degree the substructures contribute to retention, according to the model.
            }
            \label{fig:fig4}
        \end{figure}

        One of the potential strengths of the GNN models is their ability to compute saliency or gradient activation maps to better interpret e.g. retention data; i.e., to better understand what substructures contribute to the (RT) prediction. Figure \ref{fig:fig4} visualizes four different molecules from the RPLC and HILIC datasets, with gradient activation maps superimposed on the 2D structures of these molecules. A green color indicates a positive contribution to retention, while a purple color indicates a negative contribution to retention. The number of contour lines emphasizes how much the substructure contributes (either positively or negatively). Overall, as expected, non-polar substructures contribute positively to retention in RPLC, while polar substructures contribute negatively; and vice versa in HILIC. These results are expected as the non-polar stationary phase of RPLC is expected to interact with the non-polar substructures of the molecules (causing a difference in retention); and similarly, the polar stationary phase of HILIC is expected to interact with the polar substructures of the molecules. These observations largely agree with the observations of Kensert et al.\cite{intro9}, though in their study, saliency maps were used and negative contribution to retention discarded. 
    
\section{Conclusions}\label{sec:4}
    One of the main motivations behind MolGraph was to develop a GNN package highly compatible with TF and Keras. Consequently, MolGraph’s GNN layers can be built and utilized with ease — using the \code{tf.data.Dataset} API for efficient data pipelines, and the \code{keras.Sequential} or \code{keras.Functional} API for concise and flexible implementations of GNN models. For now, MolGraph limits its scope to property predictions (also known as quantative structure-property relation (QSPR) modeling); however, in prospect, it would be interesting to implement modules for more advanced GNN models, including generative and self-supervised models\cite{other32, other33}. Another possible direction is to focus on larger graphs, such as proteins and oligonucleotides, which are of great interest in pharmaceutical science\cite{other34, other35}.

\subsection*{Availability}
    Source code and license can be found at \url{https://github.com/akensert/molgraph}. Documentation and tutorials can be found at \url{https://molgraph.readthedocs.io/en/latest/}, which covers all the sections of this manuscript. Latest version of MolGraph at time of writing the manuscript: v0.6.0. 
    
\subsection*{Authors' contributions}
    AK wrote the source code and documentation for the MolGraph package. AK performed the experiments and wrote the original draft of the manuscript; GD and DC reviewed and edited the manuscript. GD and DC supervised, and acquired funding for, the project. AK, GD and DC conceptualized the project. 
    
\subsection*{Acknowledgement}
    Alexander Kensert is funded by a joint-initiative of the Research Foundation Flanders (FWO) and the Walloon Fund for Scientific Research (FNRS) (EOS – research project “Chimic” (EOS ID: 30897864)).
    
\subsection*{Conflict of interests}
    The authors declare that there is no conflict of interest.

\bibliographystyle{unsrt}  
\bibliography{references}

\begin{thebibliography}{10}

\bibitem{intro1}
Qiuji Cui, Shuai Lu, Bingwei Ni, Xian Zeng, Ying Tan, Ya~Dong Chen, and
  Hongping Zhao.
\newblock Improved prediction of aqueous solubility of novel compounds by going
  deeper with deep learning.
\newblock {\em Frontiers in oncology}, 10:121, 2020.

\bibitem{intro2}
Yan-Bin Wang, Zhu-Hong You, Shan Yang, Hai-Cheng Yi, Zhan-Heng Chen, and Kai
  Zheng.
\newblock A deep learning-based method for drug-target interaction prediction
  based on long short-term memory neural network.
\newblock {\em BMC medical informatics and decision making}, 20(2):1--9, 2020.

\bibitem{intro3}
Yaxia Yuan, Fang Zheng, and Chang-Guo Zhan.
\newblock Improved prediction of blood--brain barrier permeability through
  machine learning with combined use of molecular property-based descriptors
  and fingerprints.
\newblock {\em The AAPS journal}, 20(3):1--10, 2018.

\bibitem{intro4}
Roberto Todeschini and Viviana Consonni.
\newblock {\em Handbook of molecular descriptors}.
\newblock John Wiley \& Sons, 2008.

\bibitem{intro5}
David Rogers and Mathew Hahn.
\newblock Extended-connectivity fingerprints.
\newblock {\em Journal of chemical information and modeling}, 50(5):742--754,
  2010.

\bibitem{intro6}
Steven Kearnes, Kevin McCloskey, Marc Berndl, Vijay Pande, and Patrick Riley.
\newblock Molecular graph convolutions: moving beyond fingerprints.
\newblock {\em Journal of computer-aided molecular design}, 30(8):595--608,
  2016.

\bibitem{intro7}
David~K Duvenaud, Dougal Maclaurin, Jorge Iparraguirre, Rafael Bombarell,
  Timothy Hirzel, Al{\'a}n Aspuru-Guzik, and Ryan~P Adams.
\newblock Convolutional networks on graphs for learning molecular fingerprints.
\newblock {\em Advances in neural information processing systems}, 28, 2015.

\bibitem{intro8}
Justin Gilmer, Samuel~S Schoenholz, Patrick~F Riley, Oriol Vinyals, and
  George~E Dahl.
\newblock Neural message passing for quantum chemistry.
\newblock In {\em International conference on machine learning}, pages
  1263--1272. PMLR, 2017.

\bibitem{intro9}
Alexander Kensert, Robbin Bouwmeester, Kyriakos Efthymiadis, Peter Van~Broeck,
  Gert Desmet, and Deirdre Cabooter.
\newblock Graph convolutional networks for improved prediction and
  interpretability of chromatographic retention data.
\newblock {\em Analytical Chemistry}, 93(47):15633--15641, 2021.

\bibitem{intro10}
Dejun Jiang, Zhenxing Wu, Chang-Yu Hsieh, Guangyong Chen, Ben Liao, Zhe Wang,
  Chao Shen, Dongsheng Cao, Jian Wu, and Tingjun Hou.
\newblock Could graph neural networks learn better molecular representation for
  drug discovery? a comparison study of descriptor-based and graph-based
  models.
\newblock {\em Journal of cheminformatics}, 13(1):1--23, 2021.

\bibitem{intro11}
Phillip~E Pope, Soheil Kolouri, Mohammad Rostami, Charles~E Martin, and Heiko
  Hoffmann.
\newblock Explainability methods for graph convolutional neural networks.
\newblock In {\em Proceedings of the IEEE/CVF Conference on Computer Vision and
  Pattern Recognition}, pages 10772--10781, 2019.

\bibitem{intro12}
Sanjay~Surendranath Girija.
\newblock Tensorflow: Large-scale machine learning on heterogeneous distributed
  systems.
\newblock {\em Software available from tensorflow. org}, 39(9), 2016.

\bibitem{intro13}
Francois Chollet et~al.
\newblock Keras, 2015.

\bibitem{intro14}
Adam Paszke, Sam Gross, Francisco Massa, Adam Lerer, James Bradbury, Gregory
  Chanan, Trevor Killeen, Zeming Lin, Natalia Gimelshein, Luca Antiga, et~al.
\newblock Pytorch: An imperative style, high-performance deep learning library.
\newblock {\em Advances in neural information processing systems}, 32, 2019.

\bibitem{intro15}
Daniele Grattarola and Cesare Alippi.
\newblock Graph neural networks in tensorflow and keras with spektral
  [application notes].
\newblock {\em IEEE Computational Intelligence Magazine}, 16(1):99--106, 2021.

\bibitem{intro16}
Minjie Wang, Da~Zheng, Zihao Ye, Quan Gan, Mufei Li, Xiang Song, Jinjing Zhou,
  Chao Ma, Lingfan Yu, Yu~Gai, et~al.
\newblock Deep graph library: A graph-centric, highly-performant package for
  graph neural networks.
\newblock {\em arXiv preprint arXiv:1909.01315}, 2019.

\bibitem{intro17}
Bharath Ramsundar, Peter Eastman, Patrick Walters, Vijay Pande, Karl Leswing,
  and Zhenqin Wu.
\newblock {\em Deep Learning for the Life Sciences}.
\newblock O'Reilly Media, 2019.

\bibitem{intro18}
Matthias Fey and Jan~Eric Lenssen.
\newblock Fast graph representation learning with pytorch geometric.
\newblock {\em arXiv preprint arXiv:1903.02428}, 2019.

\bibitem{intro19}
Peter~W Battaglia, Jessica~B Hamrick, Victor Bapst, Alvaro Sanchez-Gonzalez,
  Vinicius Zambaldi, Mateusz Malinowski, Andrea Tacchetti, David Raposo, Adam
  Santoro, Ryan Faulkner, et~al.
\newblock Relational inductive biases, deep learning, and graph networks.
\newblock {\em arXiv preprint arXiv:1806.01261}, 2018.

\bibitem{other20}
Kristof~T Sch{\"u}tt, Farhad Arbabzadah, Stefan Chmiela, Klaus~R M{\"u}ller,
  and Alexandre Tkatchenko.
\newblock Quantum-chemical insights from deep tensor neural networks.
\newblock {\em Nature communications}, 8(1):1--8, 2017.

\bibitem{other21}
Max Welling and Thomas~N Kipf.
\newblock Semi-supervised classification with graph convolutional networks.
\newblock In {\em J. International Conference on Learning Representations (ICLR
  2017)}, 2016.

\bibitem{other22}
Vijay~Prakash Dwivedi, Chaitanya~K Joshi, Thomas Laurent, Yoshua Bengio, and
  Xavier Bresson.
\newblock Benchmarking graph neural networks.
\newblock {\em arXiv preprint arXiv:2003.00982}, 2020.

\bibitem{other23}
Keyulu Xu, Weihua Hu, Jure Leskovec, and Stefanie Jegelka.
\newblock How powerful are graph neural networks?
\newblock {\em arXiv preprint arXiv:1810.00826}, 2018.

\bibitem{other24}
Petar Velickovic, Guillem Cucurull, Arantxa Casanova, Adriana Romero, Pietro
  Lio, and Yoshua Bengio.
\newblock Graph attention networks.
\newblock {\em stat}, 1050:20, 2017.

\bibitem{other25}
Federico Monti, Davide Boscaini, Jonathan Masci, Emanuele Rodola, Jan Svoboda,
  and Michael~M Bronstein.
\newblock Geometric deep learning on graphs and manifolds using mixture model
  cnns.
\newblock In {\em Proceedings of the IEEE conference on computer vision and
  pattern recognition}, pages 5115--5124, 2017.

\bibitem{other36}
Jacob Devlin, Ming-Wei Chang, Kenton Lee, and Kristina Toutanova.
\newblock Bert: Pre-training of deep bidirectional transformers for language
  understanding.
\newblock {\em arXiv preprint arXiv:1810.04805}, 2018.

\bibitem{other37}
Kaiming He, Xiangyu Zhang, Shaoqing Ren, and Jian Sun.
\newblock Deep residual learning for image recognition.
\newblock In {\em Proceedings of the IEEE conference on computer vision and
  pattern recognition}, pages 770--778, 2016.

\bibitem{other26}
Zhenqin Wu, Bharath Ramsundar, Evan~N Feinberg, Joseph Gomes, Caleb Geniesse,
  Aneesh~S Pappu, Karl Leswing, and Vijay Pande.
\newblock Moleculenet: a benchmark for molecular machine learning.
\newblock {\em Chemical science}, 9(2):513--530, 2018.

\bibitem{other27}
Xavier Domingo-Almenara, Carlos Guijas, Elizabeth Billings, J~Rafael
  Montenegro-Burke, Winnie Uritboonthai, Aries~E Aisporna, Emily Chen, H~Paul
  Benton, and Gary Siuzdak.
\newblock The metlin small molecule dataset for machine learning-based
  retention time prediction.
\newblock {\em Nature communications}, 10(1):1--9, 2019.

\bibitem{other28}
Paolo Bonini, Tobias Kind, Hiroshi Tsugawa, Dinesh~Kumar Barupal, and Oliver
  Fiehn.
\newblock Retip: retention time prediction for compound annotation in
  untargeted metabolomics.
\newblock {\em Analytical chemistry}, 92(11):7515--7522, 2020.

\bibitem{other29}
Diederik~P Kingma and Jimmy Ba.
\newblock Adam: A method for stochastic optimization.
\newblock {\em arXiv preprint arXiv:1412.6980}, 2014.

\bibitem{other30}
Hiroshi Tsugawa, Tobias Kind, Ryo Nakabayashi, Daichi Yukihira, Wataru Tanaka,
  Tomas Cajka, Kazuki Saito, Oliver Fiehn, and Masanori Arita.
\newblock Hydrogen rearrangement rules: computational ms/ms fragmentation and
  structure elucidation using ms-finder software.
\newblock {\em Analytical chemistry}, 88(16):7946--7958, 2016.

\bibitem{other31}
Zijuan Lai, Hiroshi Tsugawa, Gert Wohlgemuth, Sajjan Mehta, Matthew Mueller,
  Yuxuan Zheng, Atsushi Ogiwara, John Meissen, Megan Showalter, Kohei Takeuchi,
  et~al.
\newblock Identifying metabolites by integrating metabolome databases with mass
  spectrometry cheminformatics.
\newblock {\em Nature methods}, 15(1):53--56, 2018.

\bibitem{other32}
Nicola De~Cao and Thomas Kipf.
\newblock Molgan: An implicit generative model for small molecular graphs.
\newblock {\em arXiv preprint arXiv:1805.11973}, 2018.

\bibitem{other33}
Weihua Hu, Bowen Liu, Joseph Gomes, Marinka Zitnik, Percy Liang, Vijay Pande,
  and Jure Leskovec.
\newblock Strategies for pre-training graph neural networks.
\newblock {\em arXiv preprint arXiv:1905.12265}, 2019.

\bibitem{other34}
Hiroshi Tsugawa, Tobias Kind, Ryo Nakabayashi, Daichi Yukihira, Wataru Tanaka,
  Tomas Cajka, Kazuki Saito, Oliver Fiehn, and Masanori Arita.
\newblock Hydrogen rearrangement rules: computational ms/ms fragmentation and
  structure elucidation using ms-finder software.
\newblock {\em Analytical chemistry}, 88(16):7946--7958, 2016.

\bibitem{other35}
Thomas~C Roberts, Robert Langer, and Matthew~JA Wood.
\newblock Advances in oligonucleotide drug delivery.
\newblock {\em Nature Reviews Drug Discovery}, 19(10):673--694, 2020.

\end{thebibliography}

\end{document}